\documentclass[conference]{IEEEtran}
\IEEEoverridecommandlockouts
\usepackage[ruled,linesnumbered,vlined]{algorithm2e}
\usepackage{times}  
\usepackage{helvet}  
\usepackage{courier}  
\usepackage{url,amsmath,amsthm,mathrsfs,amssymb,multicol,multirow}  
\usepackage{graphicx,enumitem}  
\usepackage{bbm}
\usepackage{booktabs,tabularx,threeparttable}
\usepackage{color, xcolor}
\usepackage[normalem]{ulem}
\usepackage{makecell}
\theoremstyle{definition}
\newtheorem{definition}{Definition}
\usepackage{bm}
\usepackage{cite}
\usepackage{amsfonts}
\usepackage{textcomp}
\def\BibTeX{{\rm B\kern-.05em{\sc i\kern-.025em b}\kern-.08em
    T\kern-.1667em\lower.7ex\hbox{E}\kern-.125emX}}

\begin{document}

\title{Efficient Learning-based Graph Simulation for Temporal Graphs
\thanks{* Corresponding author: Dawei Cheng (dcheng@tongji.edu.cn)\\
This work was supported by the National Science Foundation of China (Grant no. 62472317), the Fundamental Research Funds for the Central Universities, the Shanghai Science and Technology Innovation Action Plan Project (Grant no. 22YS1400600 and 24692118300). Ying Zhang is supported by ARC LP210301046 and DP230101445. Xiaoyang Wang is supported by ARC DP230101445 and DP240101322.}
}

\author{\IEEEauthorblockN{1\textsuperscript{st} Sheng Xiang}
\IEEEauthorblockA{\textit{Australian Artificial Intelligence Institute} \\
\textit{University of Technology Sydney}\\
Sydney, Australia \\
sheng.xiang@uts.edu.au}
\and
\IEEEauthorblockN{2\textsuperscript{nd} Chenhao Xu}
\IEEEauthorblockA{\textit{School of Computer Science} \\
\textit{Peking University}\\
Beijing, China \\
xuchenhao@stu.pku.edu.cn}
\and
\IEEEauthorblockN{3\textsuperscript{rd} Dawei Cheng\textsuperscript{*}}
\IEEEauthorblockA{\textit{School of Computer Science and Technology} \\
\textit{Tongji University}\\
\textit{Shanghai Artificial Intelligence Laboratory}\\
Shanghai, China \\
dcheng@tongji.edu.cn}
\and
\IEEEauthorblockN{4\textsuperscript{th} Xiaoyang Wang}
\IEEEauthorblockA{\textit{School of Computer Science and Engineering} \\
\textit{University of New South Wales}\\
Sydney, Australia \\
xiaoyang.wang1@unsw.edu.au}
\and
\IEEEauthorblockN{5\textsuperscript{th} Ying Zhang}
\IEEEauthorblockA{\textit{Australia Artificial Intelligence Institute} \\
\textit{University of Technology Sydney}\\
Sydney, Australia \\
ying.zhang@uts.edu.au}
}

\maketitle

\begin{abstract}
Graph simulation has recently received a surge of attention in graph processing and analytics. 
In real-life applications, e.g. social science, biology, and chemistry, many graphs are composed of a series of evolving graphs (i.e., temporal graphs).
While most of the existing graph generators focus on static graphs, the temporal information of the graphs is ignored.
In this paper, we focus on simulating temporal graphs, which aim to reproduce the structural and temporal properties of the observed real-life temporal graphs. In this paper, we first give an overview of the existing temporal graph generators, including recently emerged learning-based approaches. Most of these learning-based methods suffer from one of the limitations: low efficiency in training or slow generating, especially for temporal random walk-based methods. 
Therefore, we propose an efficient learning-based approach to generate graph snapshots, namely temporal graph autoencoder (TGAE). 
Specifically, we propose an attention-based graph encoder to encode temporal and structural characteristics on sampled ego-graphs. And we proposed an ego-graph decoder that can achieve a good trade-off between simulation quality and efficiency in temporal graph generation.
Finally, the experimental evaluation is conducted among our proposed TGAE and representative temporal graph generators on real-life temporal graphs and synthesized graphs. It is reported that our proposed approach outperforms the state-of-the-art temporal graph generators by means of simulation quality and efficiency.
\end{abstract}


\section{Introduction}\label{sec:introduction}
Due to the graph’s strong expressive power, a host of researchers in fields such as e-commerce, cybersecurity, social
networks, military, public health, and many more, are turning to graph modeling to support real-world data analysis~\cite{GSOTA2020,Goyal2018Graph,Wu2019ACS,ZHOU202057,Zhang2022DeepLO,xiang2021general}.
An important line of research for graph processing and analytics is the simulation of graphs 
which is used for many purposes such as tackling the inaccessibility of the whole real-life graphs and a better understanding of the distribution of graph structures and other features~\cite{julia2020resp,1983Stochastic,Jure2005as20}.
There is a long history of study for the graph simulation in many research fields such as Database and Machine learning~\cite{xiang2021general,Ma2019GraphSO,Xie2022SelfSupervisedLO}.
Recently, many research efforts have been devoted to design advanced generative models 
which can significantly enhance the simulation quality, thanks to the recent development of deep learning techniques.
Nevertheless, we notice that most of the existing works aim to simulate the static graphs.
While in many real-life applications, the data are naturally modeled as temporal graphs (a.k.a time-evolving graphs) where the graph evolves with the time. For instance, in the online finance networks and e-commerce networks~\cite{cheng2020spatio,Zhu2019AliGraphAC}, the edges consists of a sequence of transactions with timestamps for users or products. In the location-based service networks with regarding to the Point of Interests (POIs)~\cite{Jahromi2016SimulatingHM}, an edge corresponds to a visit of an user towards a POI (e.g., a restaurant) at a particular time. 
In these applications, it is critical to capture the \textit{evolution of the graphs} over the time;
that is, the structure of the real graph will evolve with time, and hence the edge generative probability distribution of the nodes on the graph will change with the passage of time. 
For most graph simulators, which do not properly consider the temporal information of the graphs,
a straightforward way is to simulate the temporal graphs based on some particular timestamps, i.e., learn the snapshots of the evolving graph at these timestamps. Unfortunately, this is not feasible in practice because it is cost-prohibitive to learn and simulate many snapshots of the graphs.
Thus, to effectively and efficiently support temporal graph simulation in many key applications such as the generation of new drug molecules~\cite{jin2018junction}, chemical reaction pathway simulation~\cite{McDermott2021AGN}, router load balancing~\cite{Jure2005as20}, and pandemic trajectory generation~\cite{Gonzlez2008UnderstandingIH}, 
it is essential to develop advanced graph generative models which can capture both structure and time properties of the time-evolving graphs.

\begin{figure}[t!]
    \centering
    \includegraphics[width=1\linewidth]{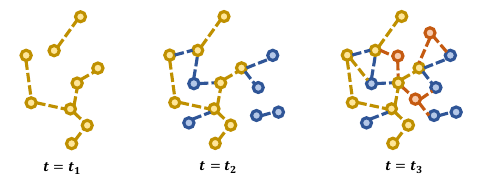}
    \caption{An example of time-evolving graph.}
    \label{fig:intro}
\end{figure}

\vspace{1mm}
\noindent\textbf{Motivation.}
\label{subsec:motivation}
We can store the time-evolving graph as a collection of graph snapshots (i.e. a series of time-stamped static graphs). The collection of graph snapshots contains all edges, nodes, and their corresponding timestamps. Specifically, as shown in Figure~\ref{fig:intro}, after a period of time, several temporal nodes and edges are added to this time-stamped graph snapshot. The traditional way to model such a time-evolving graph is to aggregate timestamps into a series of snapshots. There is also an alternative way for dynamic temporal edges, i.e. temporal random walks, to model a time-evolving graph. Some recently developed learning-based temporal graph generative methods~\cite{zhou2020data,zhang2021tggan,Gupta_Manchanda_Bedathur_Ranu_2022_TIGGER}, namely \textit{TagGen} and its successors, reconstructs a set of temporal random walks to assemble a synthetic graph. However, a disadvantage comes from the inevitable bias of decomposing the time-evolving graphs into a set of temporal random walks. In this case, we have to live with the extra time and space required by a lot of operations of random walk sampling. If the number of walks is too large, the large number of training samples will bring unbearable computational costs when training deep generative models; if the number of walks is too small, the rich temporal structure properties may be lost during model training. 
The other disadvantage comes from the $O(T^2)$ factor in the time and space complexity analysis of \textit{TagGen} where $T$ is the number of distinct timestamps.
complexity of time consumption and memory usage. This directly limits the efficiency and the scalability of the graph simulation model, especially when we need to simulate fine-grained time-evolving graphs. 

This motives us to develop a new generative model for time-evolving graphs with better efficiency (scalability) and simulation quality.
Particularly, to better capture both the structure and temporal information of the graph,
we design a new temporal ego-graph based sampling approach for the temporal graph and we re-weight the input temporal nodes according to the degree of the node, so as to give priority to learning the local temporal structure of the representative nodes. 
For these representative nodes, we sample their ego-graphs to learn the representative temporal graph structure. We encode and reconstruct the ego-graphs' structure through the temporal graph attention mechanism. As to ego-graphs with $k$ radius, we stack $k$ temporal graph attention layers to layer by layer transmit messages from the periphery nodes to the center nodes. After that, we use the cross-entropy loss to learn directly from the representative temporal graph structure.
We also designed a GPU-friendly parallel temporal ego-graph training strategy and the corresponding approximate objective function to reduce the number of steps of model training and achieve efficient model training. Specifically, we combine all the sampled ego-graphs to assemble $k$ bipartite computation graphs to parallelize the computation process and reduce redundant repeats.
After our improvements, the number of steps of model training is $O(\frac{nT}{n_s})$, and the space consumption of model training is $O(n\times(T+n_s))$, where $n_s$ denotes the hyper-parameter of the number of sampled initial nodes and $n$ is the number of graph nodes.
As demonstrated in the empirical study, our new approach can achieve much better simulation quality compared to the state-of-the-art technique.
Moreover, the new approach also win out from space consumption, efficiency and scalability perspectives. 

\vspace{1mm}
\noindent\textbf{Contribution.}
\label{subsec:contribution}
Our principal contributions in this paper are summarized as follows:

\begin{itemize}
\item We propose a new ego-graph based sampling strategy for the temporal graph, which has a stronger expressiveness to better capture the local structure 
and temporal properties of the temporal graph. We also design an efficient ego-graph sampling strategy and GPU-friendly parallelization ego-graph training strategy, as well as approximate optimization objectives, to achieve scalable model training. 

\item For better simulation of temporal graphs, we design a Temporal Graph Autoencoder (TGAE). Specifically, temporal node messages are passed from peripheral nodes to the central node in the ego-graph through a temporal graph attention mechanism. Then, the entire ego-graph is reconstructed evolutionarily from the central node through a variational autoencoder.

\item Through the extensive experiments on both real-life and synthetic time-evolving graphs, we boast that our new approach significantly outperforms
the state-of-the-art in terms of temporal graph simulation quality, efficiency, scalability and space consumption.
\end{itemize}

\vspace{1mm}
\noindent\textbf{Roadmap}.
The rest of the paper is organized as follows. 
In Section~\ref{sec:preliminary}, we formally define the problem and provide the related works of this paper. We then present the details of our proposed graph generator and introduce the optimizing targets in Section~\ref{sec:model}.
Comprehensive experimental results for temporal graph generators are presented in Section~\ref{sec:exprm}. 
Section~\ref{sec:conclusion} concludes the paper.

\section{Related Works}
This section presents a review of recent literature on temporal graph learning and graph generative models.

\subsection{Temporal Graph Learning}
Recently, a large number of work has appeared in temporal graph learning. In these works, temporal graphs can be represented by a set of timestamped nodes and edges. For instance, Spatio-temporal graph convolution networks are leveraged on traffic forecasting~\cite{yu2017spatio}. The linkage evolution process is leveraged on temporal graph representation learning~\cite{li2018deep}. The graph attention mechanism is proposed to focus on the crucial part of the graph data for time-evolving graph learning~\cite{liu2019towards}. Dynamic parameters of graph convolution networks are proposed to improve the temporal graph embeddings~\cite{Pareja2020EvolveGCNEG}. \cite{wang2021apan} decouples model inference and graph computation to alleviate the damage of the heavy graph query operation to the speed of model inference. \cite{yang2021discrete} proposes the hyperbolic temporal graph network (HTGN) that fully takes advantage of the exponential capacity and hierarchical awareness of hyperbolic geometry. However, separately focusing on temporal property and topological structure limits the embedding quality. To address this issue, our encoder aims to encode the temporal and topological structure together into one embedding. In this paper, we propose to use temporal graph attention as the encoder component of our proposed temporal graph auto-encoder (TGAE).

\subsection{Graph Generative Model}
Early graph simulation problems were often solved by models for static graphs. For instances, random graph model~\cite{erdHos1960evolution,chakrabarti2004r}, small world graph~\cite{Watts1998Collective,grabow2015collective}, preferential attachment~\cite{akoglu2009rtg, albert2002statistical, kleinberg1999web}, stochastic blockmodels~\cite{1983Stochastic,2011Stochastic,airoldi2008mixed}. To improve the quality of graph simulation, many deep graph generative models have recently been proposed to learn graph generative distributions directly from observed graphs. For instances, SBMGNN~\cite{mehta2019stochastic} uses a graph neural network to parameterize the overlapping stochastic blockmodels, GraphRNN~\cite{you2018graphrnn} uses a recurrent neural network to learn to draw a graph from scratch, and NetGAN~\cite{bojchevski2018netgan} uses a generative adversarial network to learn the generative distribution of random walks. However, these works ignore the dynamic nature of real-world graphs, i.e., the topology structure of a graph evolves over time. To solve this problem, for example, TagGen~\cite{zhou2020data} and TGGAN~\cite{zhang2021tggan} were recently proposed to use generative adversarial networks (GANs) to simulate temporal graphs, and make learning-based methods achieved the best temporal graph simulation quality. Recently, there has been a work on temporal graph simulation~\cite{Ma2019GraphSO}, which is essentially different from our study, as they leverage simulation on temporal graph pattern matching. Different from generating random walks, our method directly learns the generative distribution of observed temporal graphs. We also proposed an efficient learning strategy that achieves a good trade-off between simulating quality and efficiency.

\subsection{Temporal Graph Generator}
\label{subsec:state}
Several methods have been proposed for temporal graph generation. The Motif Transition Model~\cite{liu2023using} models the transition process of dynamic motifs in graphs, providing a simple and scalable simulator for dynamic graphs. Similarly, RTGEN++~\cite{massri2023rtgen++} models the evolution process of degree distributions over time, also achieving high scalability. Temporal Edge Distribution (TED)~\cite{zheng2024temporal} model addresses the challenge of considering temporal community attributes for generating temporal graphs.

However, compared with non-learning-based methods, learning-based methods have significantly improved the generation quality of temporal graphs. For example, TagGen~\cite{zhou2020data} employs a temporal random walk-based approach to capture temporal dependencies by generating new temporal random walks based on those sampled from observed temporal graphs. Building upon TagGen, TGGAN~\cite{zhang2021tggan} introduces a generative adversarial network framework, where a generator synthesizes new temporal random walks and a discriminator evaluates their validity. In addition, DYMOND~\cite{WWW21DYMOND} models dynamic motifs on graphs using a parameterized model, aiming to capture temporal motif distributions accurately. However, these methods face limitations in computational complexity. For example, the generation process of TagGen and TGGAN has a complexity of $O(n^2\times T^2)$, and the training and inference time complexity of DYMOND is $O(n^3\times T)$. TIGGER~\cite{Gupta_Manchanda_Bedathur_Ranu_2022_TIGGER} further improves the generation quality and scalability based on TagGen, reducing the complexity to $O(n\times M)$, where $M$ is the number of edges. However, TIGGER's generation model is limited by the inherent constraints of random walk-based methods, requiring a large number of random walks to achieve high-quality generation.

Unlike generating random walks, our proposed Temporal Graph Auto-Encoder (TGAE) directly learns the generation distribution of the observed temporal graph, with a computational complexity of $O(n^2\times T)$. We also propose an efficient parallel learning strategy, achieving a good balance between simulation quality and efficiency. After parallelization, the computational steps of our method can be reduced to $O(n\times T)$.

\section{Preliminary}
\label{sec:preliminary}


Table~\ref{tb:symbols} summarizes the symbols introduced in this paper. In this section, we formalize the graph generation problem in the temporal graph~\cite{paranjape2017motifs,zhou2020data}. Given a temporal graph $\Tilde{G}$, we model the temporal graph as a series of graph snapshots $\{G^{t_1},...,G^{T}\}$, which include temporal nodes $\{v_1^{t_{v_1}},...,v_n^{t_{v_n}}\}$ and temporal edges $\{e_1^{t_{e_1}},...,e_m^{t_{e_m}}\}$.
The definitions of temporal nodes and edges are as follows:

\begin{table}\vspace{-0pt}
  \caption{The summary of symbols}\label{tb:symbols}\vspace{-0pt}
  \centering
  \begin{tabular}{|c|l|}
     \hline
     \textbf{Symbol} & \textbf{Definition} \\ \hline \hline
     $\mathbf{X}$ & the input features of node occurrences\\ \hline
     $\mathbf{\hat{Y}}$ & the probability of generating an edge\\ \hline
     $n$ & the total number of nodes \\ \hline
     $m$ & the total number of edges \\ \hline
     $T$ & the number of timestamps in $\Tilde{G}$\\ \hline
     $\Tilde{V}=\{v_1^{t_{v_1}},...,v_n^{t_{v_n}}\}$ & the set of temporal nodes\\ \hline
     $\Tilde{E}=\{e_1^{t_{e_1}},...,e_m^{t_{e_m}}\}$ & the set of temporal edges\\ \hline
     $\mathbf{N}(\cdot)$ & the neighborhood function\\ \hline
     $d$ & the number of dimension\\ \hline
  \end{tabular}
  \vspace{-0pt}
\end{table}

\begin{definition}
    \label{Temporal Nodes and Edges}
    \textbf{Temporal Nodes and Edges.} \textit{In a temporal graph, a node $v_i$ is associated with a node id $i$ and timestamps $v_i$ occurred with $v_i=\{v_i^{t_1},v_i^{t_2},...\}$. Same as temporal node $v$, an edge $e_j$ is associated with a timestamp $t_{e_j}$ and two temporal nodes $u^{t_{e_j}}$ and $v^{t_{e_j}}$. For temporal nodes and edges in the same timestamp $t$, the set of nodes and edges are defined as $V^t$ and $E^t$, respectively.}
\end{definition}

We choose to model the temporal graph as a series of graph snapshots for two main reasons. First, capturing the state of the graph at discrete time intervals aligns well with the periodic nature of some dynamic graphs. Second, this representation simplifies the temporal dynamics into a series of static graphs, making it more tractable for certain general graph generative methods, e.g., VGAE~\cite{kipf2016variational} and NetGAN~\cite{bojchevski2018netgan}. While previous temporal graph generation work~\cite{zhou2020data} describes the temporal graph as a set composed of timestamped edges and nodes, we acknowledge that such representations can provide a more granular view of the graph's evolution. Our methodology is adaptable and can support this representation as well. If a graph is provided in this format, our approach can be extended to process and generate graphs that reflect the temporal changes among all time stamps.

Given the sets of temporal nodes $\Tilde{V}$ and temporal edges $\Tilde{E}$, the definitions of temporal graph and graph snapshots are as follows:

\begin{definition}
    \label{Temporal Graph}
    \textbf{Temporal Graph.} \textit{A temporal graph $\Tilde{G}=\{G^{t_1},...,G^{T}\}$ is formed by a series of temporal graph snapshots $G^{t}=(V^{t}, E^{t})$ with $t=1...T$. And a snapshot $G^{t}$ is associated with a timestamp $t$, temporal nodes $V^t=\{v_1^t,v_2^t,...\}$, and temporal edges $E^t=\{e_1^t,e_2^t,...\}$.}
\end{definition}

In existing graph generators~\cite{Rendsburg2020NetGANWG,bojchevski2018netgan,you2018graphrnn}, the graph neighborhood $\mathbf{N}(v)$ of node $v$ is defined as static one. Here, we generalize the definition of graph neighborhood to the temporal graph, which is defined as follows:

\begin{definition}
    \label{Temporal Neighborhood}
    \textbf{Temporal Neighborhood.} \textit{Given a temporal node $v^t$, the neighborhood of $v^t$ is defined as $\mathbf{N}(v^t)=\{v_i^t|f_{sp}(u_i^t,v^t)\leq d_{\mathbf{N}},|t_v-t_{u_i^t}|\leq t_{\mathbf{N}}\}$, where $f_{sp}(\cdot|\cdot)$ denotes the shortest path length between two nodes, $d_{\mathbf{N}}$ and $t_{\mathbf{N}}$ denote the path length and time window length, respectively, which are the hyper-parameters.}
\end{definition}

To represent the neighborhood in temporal graph, the previous work defines $k$-length temporal walks~\cite{zhou2020data,nguyen2018continuous}. Different from them, we relax this definition to the ego graph with a radius of $k$, so that we can capture the local temporal neighbor structure of the observed graph.
We consider all the neighbor nodes within the time window of the graph and take these nodes and their corresponding temporal edges as $k$-radius temporal ego-graph, which is defined as follows:

\begin{definition}
    \label{k-Radius Temporal Ego-Graph}
    \textbf{k-Radius Temporal Ego-Graph.} \textit{Given a temporal node $v^t$, a $k$-radius temporal ego-graph $\Tilde{G}_{ego}(v^t)$ is composed of temporal nodes $\Tilde{V}_{ego}(v^t)$ and edges $\Tilde{E}_{ego}(v^t)$ corresponding to the temporal neighborhood of $v$, i.e., $\mathbf{N}(v^t)$. }%
\end{definition}

\vspace{2mm}
\noindent \textbf{Problem statement.} 
In this paper, we focus on the graph simulation task for directed temporal graphs. The temporal graph generator, i.e., the proposed method for graph simulation takes the topology structure with/w.o. node features as input.  We formally define the temporal graph generation problem as follows:

\noindent \textbf{Input:} \textit{We use a temporal graph $\Tilde{G}=\{\Tilde{V},\Tilde{E}\}$ as the input of temporal graph generator for simulation}.

\noindent \textbf{Output:} \textit{The generated temporal graph $\Tilde{G}'=\{\Tilde{V}',\Tilde{E}'\}$ with highly preserved both structural and temporal properties from observed temporal graph}.

\begin{figure*}[t!]\vspace{-0pt}
    \centering
    \includegraphics[width=1\linewidth]{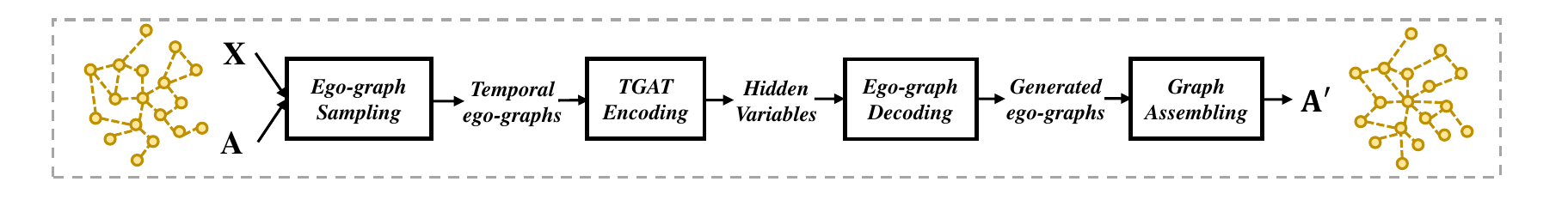}\vspace{-0pt}
    \caption{The framework of our proposed TGAE. TGAE is composed of four parts: (1) ego-graph sampling, (2) temporal graph encoding, (3) ego-graph decoding, and (4) temporal graph assembling.}
    \label{fig:architec}\vspace{-0pt}
\end{figure*}

\vspace{2mm}
\noindent \textbf{Evaluation of preserving temporal structure.}
The key of the temporal graph simulation is to keep the structure information along the whole involving procedure.
Ideally, for every timestamp, the observed time-involving graph and the simulated corresponding graph have the same distribution.
To verify the performance of preserving the temporal structure, we use the following two evaluation approaches for the temporal graph generative algorithms: 

\noindent (1) calculating the difference of graph statistics for the snapshots under the same timestamp. Specifically, we randomly choose a timestamp and accumulate the nodes and edges generated from the initial timestamp to the current timestamp to get the generated graph snapshot. Similarly, the snapshot of the original graph can be obtained by accumulating temporal edges and nodes in the same way. The difference between the generated snapshot and the original snapshot can be obtained by comparing the differences in graph statistics between the two. A distinct description of each graph statistic is introduced in Table~\ref{tab:metrics} of Section~\ref{sec:exprm}.

\noindent (2) comparing the difference in temporal motif distribution. Specifically, after generating the last timestamp, we get the snapshot of the whole temporal graph. By computing the temporal motif distribution in graph snapshot (e.g., via counting the 3-edge temporal motifs~\cite{paranjape2017motifs}), we can get the motif distribution of the generated graph and the original graph. Then we use a Gaussian kernel of the total variation (TV) to measure the distance between the two motif distributions. After that, we use the popular evaluating metric Maximum Mean Discrepancy (MMD) to measure the similarity of two distributions. Assuming that $p$ and $q$ denote the original graph's motif distribution and the generated graph's, respectively, the measurement of two distribution is formulated as follows: 
\begin{equation}
    \begin{split}
        \text{TV}(p,q)=&\mathbb{E}_{i}[||\pi_p(i)-\pi_q(i)||]\\
        \text{MMD}^2(p||q)=&\mathbb{E}_{x,y\sim p}[k(\text{TV}(x,y))]+\mathbb{E}_{x,y\sim q}[k(\text{TV}(x,y))]\\
        -&2\mathbb{E}_{x\sim p,y\sim q}[k(\text{TV}(x,y))].
    \end{split}
\end{equation}
where $k$ denotes the Gaussian kernel and $\pi_p(i)$ denotes probability of the $i$-th motif in temporal graph distribution $p$.

\section{Our Approach}
\label{sec:model}

In this section, we introduce the detailed implementation of our proposed Temporal Graph Auto-Encoders (TGAE), including ego-graph sampling, temporal graph encoding, and ego-graph decoding, and show how our TGAE generate a new temporal graph from observed temporal graphs.

\subsection{Model Architecture}
As illustrated in Figure~\ref{fig:architec}, our proposed TGAE is composed of four parts: ego-graph sampling, temporal graph encoding, ego-graph decoding, and temporal graph assembling. Given a temporal graph $\Tilde{G}=\{G^{t_1},...,G^{T}\}$, we first convert all snapshots into one temporal adjacency matrix $\mathbf{A}_{t=1:T}\in\{1,0\}^{T\times n\times n}$. Then we sample temporal $k$-radius ego-graphs and truncate the width and depth of sampled subgraphs for efficient model training. After that, we leverage temporal graph attention networks (TGAT) on neighbor-level temporal structures to obtain hidden variables for each temporal node. After temporal graph encoding, the ego-graph decoding module generates an ego-graph corresponding to each temporal node's hidden variables. Finally, we use the generated edge probabilities to assemble a temporal graph score matrix $\mathbf{S}_{t=1:T}\in\mathbb{R}^{T\times n\times n}$. Given a temporal graph score matrix, we sample edges to generate new temporal graphs.

\begin{algorithm}[t!]
\caption{Sampling k-Radius Temporal Ego-Graph}
\label{alg:sampling_egog}
\SetKwFunction{Nodesamp}{NodeSampling}
\SetKwFunction{Egograph}{k-EgoGraph}
\SetKwFunction{Datapre}{EgoGraphDataLoader}
\SetKwProg{Fn}{Function}{}{}
\Fn{\Nodesamp{nodeset, threshold}}{
    Nodes$\gets\emptyset$;
    $i$, $u\gets 0$\;
    \eIf{length(nodeset)$\leq$ threshold}{
        \Return{nodeset}\;
    }{
        \ForEach{$i\in 1:\text{threshold}$}
        {
            $u\gets$ random.choice(nodeset)\;
            Nodes.insert($u$)\;
        }
        \Return{Nodes}\;
    }
}
\Fn{\Egograph{$\Tilde{G}$, $v^t$, $k$, $th$}}{
    ego, nodeset $\gets\emptyset$\;
    \eIf{k $\ne$ 1}{
        nodeset $\gets$ \Nodesamp{$\mathbf{N}(v^t)$, $th$}\;
        \ForEach{$u^t\in$ nodeset}{
            ego $\gets$ \Egograph{$\Tilde{G}$, $u^t$, $k-1$}\;
            nodeset.insert(ego.nodes)\;
        }
        ego $\gets$ $\Tilde{G}$.subgraph(nodeset)\;
        \Return{ego}\;
    }
    {
        nodeset $\gets$ \Nodesamp{$\mathbf{N}(v^t)$, $th$}\;
        ego $\gets$ $\Tilde{G}$.subgraph(nodeset)\;
        \Return{ego}\;
    }
}
\Fn{\Datapre{$\Tilde{G}$, $\mathbf{X}_{t=1:T}$, $k$}}{
    EgoGraphs, Nodefeatures $\gets\emptyset$\;
    \ForEach{$i\in 1:T$}{
        $\mathbf{X}^{(i)}\gets \mathbf{X}_{t=1:T}((n*(i-1)+1:n*i),:)$\;
        ego, nodefeat $\gets\emptyset$\;
        \ForEach{$v\in 1:n$}{
            ego $\gets$ \Egograph{$\Tilde{G}$, $v^i$, $k$}\;
            EgoGraphs.insert(ego)\;
            nodefeat $\gets\mathbf{X}^{(i)}(\text{ego.nodes},:)$\;
            Nodefeatures.insert(nodefeat)\;
        }
    }
    \Return{EgoGraphs, NodeFeatures}
}
\end{algorithm}

\subsection{Ego-Graph Sampling}

To extract local temporal structure, we leverage ego-graph sampling for temporal graphs. Given the temporal graph adjacency matrix $\mathbf{A}_{t=1:T}$, we first separately load each snapshot's node features $\textbf{X}^{(t)}\in\mathbb{R}^{n\times d_{in}}$ corresponding to its timestamp $t$. Then, algorithm~\ref{alg:sampling_egog} shows the procedure of sampling a $k$-radius ego-graph for each temporal node $u^t$. 
Specifically, we first choose the representative temporal node as the central node of each ego-graph. 
Then we recursively sample the neighbor nodes, where the new nodes are sampled from the neighbors of the early sampled nodes.
To reduce the training time consumption of our proposed model, we truncate the number of neighbors of nodes whose degrees exceed the threshold to achieve the trade-off between efficiency and effectiveness. This approach prevents the neighborhood graphs from exploding in size, even in dense regions of the graph.
Specifically, when we sample the neighbors of the nodes, we sample several times with replacement and get a limited number of nodes as the neighbors of the node in the ego-graph. Algorithm~\ref{alg:sampling_egog} shows the detailed procedure of preparing the ego-graphs and node features as input data for our learning-based model. Note that our proposed method set the node identity numbers as default node features.

\vspace{2mm}
\noindent\textbf{Initial Temporal Node Sampling.}
To model a complete temporal graph structure, we propose a strategy to select representative temporal nodes. 
A naive approach is to sample all temporal nodes according to a uniform distribution, however, this strategy tends to learn to generate unimportant edges~\cite{SalhaGalvan2021FastGAESG}. 
To focus on important edges to generate a high-quality temporal graph, we propose to use the probability distribution based on temporal node degree as the initial temporal node sampling strategy. 
The sampling strategy is formulated as follows:

\begin{equation}
    \begin{split}
        P(u^t)=\frac{deg_{u^t}}{\sum_{v^t\in \Tilde{V}}deg_{v^t}}
    \end{split}
\end{equation}
where $deg_{u^t}$ denotes the degree of temporal node $u^t$, i.e., the temporal neighbors associated with $u^t$. Assuming that in each epoch we sample $n_s$ temporal nodes as initial temporal nodes, we sample $n_s$ ego-graphs as the input of our encoding process. The set of initial nodes is represented as $\Tilde{V}_s$.

Unlike in random walk-based work, we reweight temporal nodes by their temporal degrees (i.e., the number of first-order temporal neighbors) to efficiently simulate high-quality temporal graphs. This re-weighting will allow our model to preferentially learn to generate neighbors of key temporal nodes. Besides, the neighbors of non-critical nodes contain a higher proportion of outlier points. Therefore, our initial node sampling strategy reduces the effect of outliers, resulting in efficient and effective model training with hardly sacrificing simulating quality.

\begin{figure}[t!]\vspace{-0pt}
    \centering    \includegraphics[width=1\linewidth]{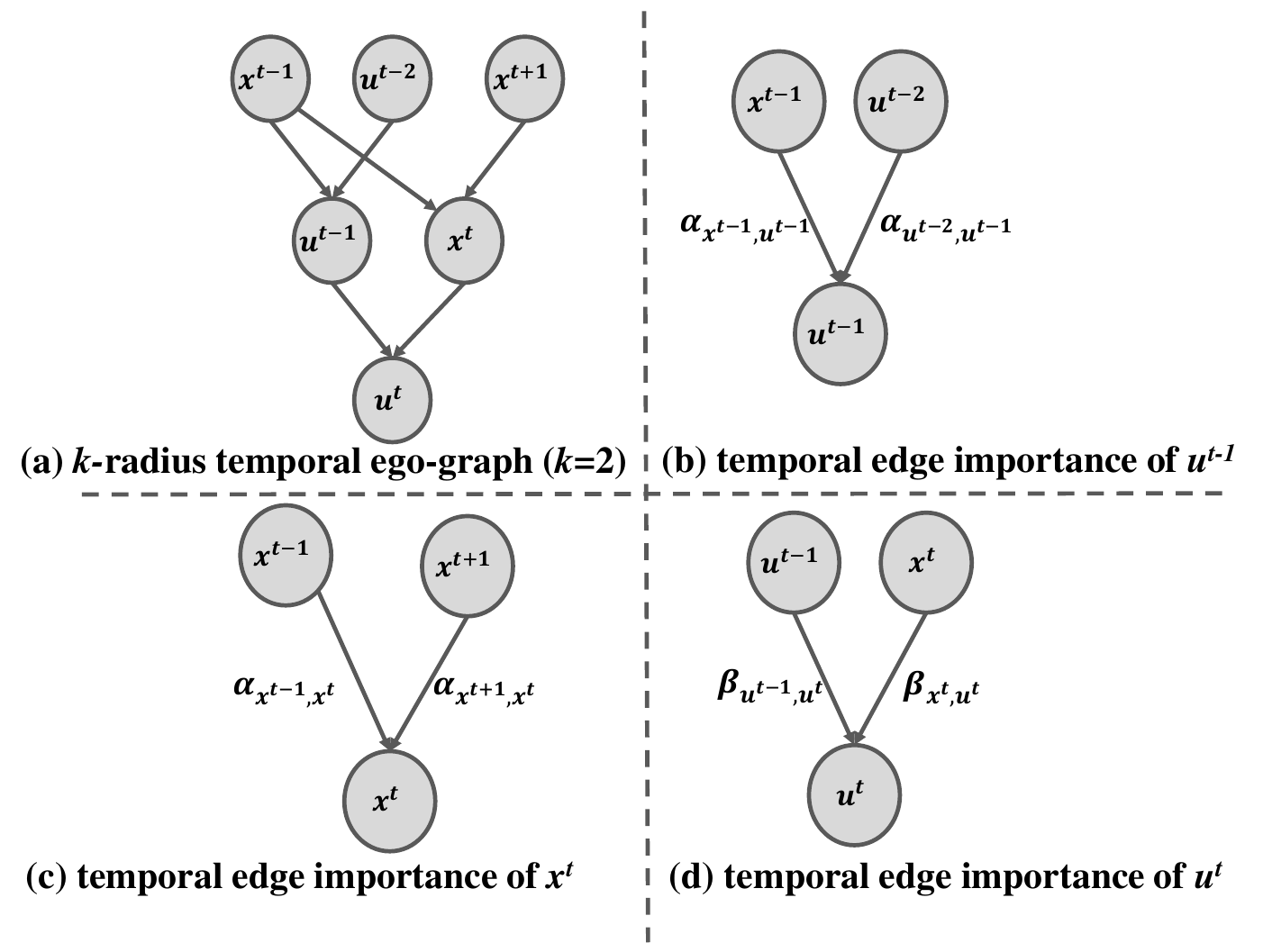}\vspace{-0pt}
    \caption{The illustration of the $k$-radius temporal ego-graph. The upper left part shows the ego-graph with the center temporal node $u^t$. The other three parts shows the edge importances calculated by $k$ stacked temporal graph attention (TGAT) layers.
}
    \label{fig:ego}\vspace{-0pt}
\end{figure}

\subsection{Temporal Graph Attention Encoding}
\label{sec:tga}
Given the temporal graph $\Tilde{G}_{ego}(v^t)=(\Tilde{V}_{ego}(v^t),\Tilde{E}_{ego}(v^t))$ and temporal node features $\textbf{X}_{ego}\in\mathbb{R}^{n_{ego}\times d_{in}}$, where $n_{ego}$ denotes the node number of ego graph and $d_{in}$ denotes the dimension of input features, we propose to employ temporal graph attention mechanism on our sampled ego graphs.
In particular, we obtain the hidden variables of the center node $u^t$ of the ego graph through leveraging temporal attention mechanism to aggregate messages from graph structures and temporal neighbors, where $d_{enc}$ denotes the dimension of hidden variables after encoding process. For each temporal ego graph, the message aggregating is formulated as follows:

\begin{equation}
    \begin{split}
        \mathbf{h}_{u^t}=&\mathbf{TGAT}_{enc}(\mathbf{X}_{ego}|\Tilde{V}_{ego}(u^t),\Tilde{E}_{ego}(u^t))\\
        =&\text{Concat}(\text{TgaHead}_1,...,\text{TgaHead}_{h_{tga}})\textbf{W}_{o}\\
    \end{split}
\end{equation}
where $\mathbf{h}_{u^t}\in \mathbb{R}^{1\times d_{att}}$ denotes one row of hidden variables of the temporal graph attention encoding layer, i.e., hidden variables on temporal node $u^t$, and $\textbf{W}_{o}\in \mathbb{R}^{h_{tga}d_{enc}\times d_{att}}$ denotes the output projection matrix, $h_{tga}$ denotes the number of heads, $d_{att}$ is the dimension of attention vector, and each head of temporal graph attention layer $\text{TgaHead}_i\in\mathbb{R}^{1\times d_{enc}}$ is formulated as follows:

\begin{equation}
    \begin{split}
        \text{TgaHead}_i=\sigma(\sum_{v^t\in \mathbf{N}(u^t)}\alpha_{u^t,v^t}^i\mathbf{h}_{u^t})\\
    \end{split}
\end{equation}
where $\sigma$ denotes the activation function and $\alpha_{u^t,v^t}^i$ denotes the importance of temporal edge $(u^t,v^t)$ in $i$-th head, which is formulated as follows:

\begin{equation}
    \begin{split}
        \alpha_{{u}^t,{v}^t}^i=\frac{\text{exp}(\text{LeakyReLU}(\mathbf{a}_{i}^T[\mathbf{h}_{u^t}||\mathbf{h}_{v^t}]))}{\sum_{{k}^t\in \mathbf{N}({v}^t)}\text{exp}(\text{LeakyReLU}(\mathbf{a}_{i}^T[\mathbf{h}_{k^t}||\mathbf{h}_{v^t}]))} \\
    \end{split}
\end{equation}
where $\mathbf{a}_{i}\in\mathbb{R}^{2d_{enc}}$ denotes the attention vector of the $i$-th attention head, and LeakyReLU denotes the non-linear activation function with a negative input slope $\alpha=0.2$.

From the example in Figure~\ref{fig:ego}, we can intuitively understand our temporal node coding process: (1) the input of the encoding process is the sampled ego-graph, which is shown in Figure~\ref{fig:ego}~(a), where we assume that the value of $k$ is 2;
(2) the first TGAT layer calculates the importance $\alpha$ of second-order neighbors (such as $x^{t-1}$ and $u^{t-2}$ in Figure~\ref{fig:ego}~(b)) to first-order neighbors (such as $u^{t-1}$ in Figure~\ref{fig:ego}~(b)); (3) the second TGAT layer calculates the importance $\beta$ of first-order neighbors to the central node $u^t$; (4) in Figure~\ref{fig:ego}~(d), the central node $u^t$ outputs the representation of this ego-graph. 
In the actual model training, we added self-loops to all temporal nodes to pass messages to themselves.

\begin{figure}[t!]\vspace{-0pt}
    \centering
    \includegraphics[width=0.9\linewidth]{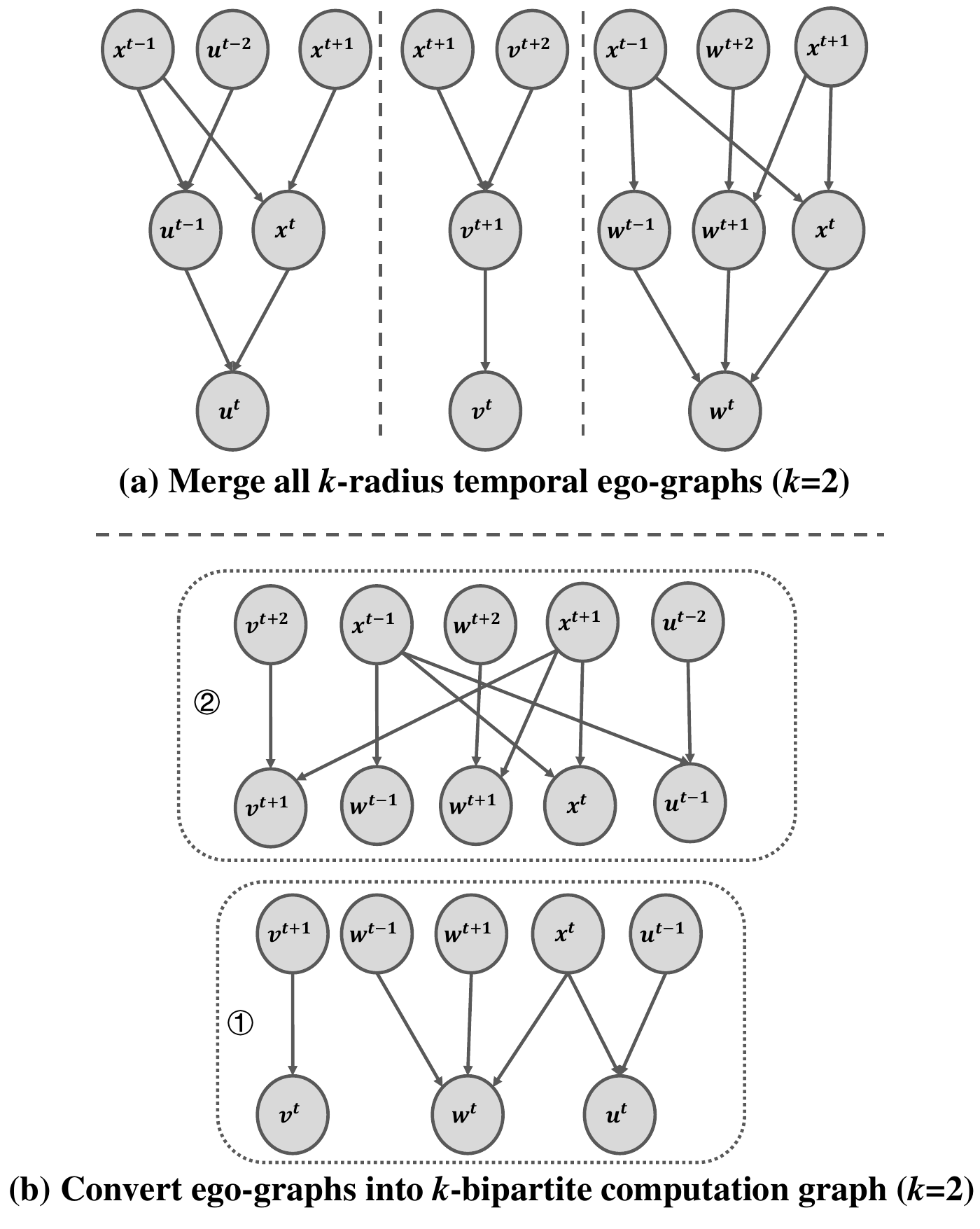}\vspace{-0pt}
    \caption{The illustration of the $k$-bipartite computation graphs. The upper part shows the initial $k$-radius temporal ego-graphs. The lower part shows the $k$-bipartite computation graphs, which are used for model training. In each bipartite computation graph, the results of the target nodes can be computed concurrently.
}
    \label{fig:bipartite}\vspace{-0pt}
\end{figure}

\vspace{2mm}
\noindent\textbf{Parallel Ego-graph Training.}
To reduce the time consumption of the encoding process, we combine multiple ego-graphs for parallel node encoding to reduce computation steps from $O(nT)$ to $O(\frac{nT}{b})$, where $b$ denotes the parallel number of temporal ego-graphs, i.e., batch size. For efficient training, we set the batch size as the size of initial sampled center node set with $b=|\Tilde{V}_s|=n_s$. Therefore, the computation step is parallelized into $O(\frac{nT}{n_s})$.
As shown in Figure~\ref{fig:bipartite}, we put all the ego-graphs together and generate $k$-bipartite graphs by vertical splitting. These bipartite graphs represent a set of temporal ego-graph neighbors of order $1$ to $k$. Specifically, we first use $S_0$ to represent the center node set of the temporal ego-graphs. Then, we use ${S_1,...,S_k}$ to respectively accommodate the $k$-order neighbors of the center nodes of these temporal ego-graphs. After that, we index the source nodes of the bipartite graph in $S_{k}$ and the target nodes of the bipartite graph in $S_{k-1}$. After getting the source and target, we get the $k$-bipartite computation graphs.
We stack $k$ TGAT layers to achieve message passing on the $k$-bipartite computation graph, and finally, get the representation of the central temporal node.

To further reduce the space consumption, we use a truncation mechanism to control space usage and ignore repeated nodes each time a new node is inserted into $S_k$.
In Algorithm~\ref{alg:sampling_egog}, to control the worst-case space requirement, we use $th$ as the threshold.
Once the total number of neighbors of a temporal node exceeds $th$, the algorithm converts from all neighbor sampling strategy to $th$-neighbor sampling strategy, and merge all the temporal ego-graphs into $k$-bipartite computation graphs for the sampled neighbor nodes instead of all the neighbor nodes.

\subsection{Ego-Graph Decoding}
To reconstruct local temporal structure distribution, we leverage ego-graph decoding process to infer the probablistic generative model for each temporal node. Given the hidden variables of temporal node $\mathbf{h}_{u^t}$, we first use two Multi-Layer Perceptrons (MLP) to infer the parameters $\mu$ and $\sigma$ of the prior distribution $\mathcal{N}(\mu,\sigma^2)$. Then algorithm~\ref{alg:decoding_egog} shows the procedure of decoding the edge probabilities of a $k$-radius ego-graph for each temporal node $u^t$. The detailed graph generation process from edge probability will be introduced in the subsection~\ref{subsec:assem}. In summary, the decoding process has a space complexity of $O(n\times(T+n_s))$, where $n_s$ denotes the number of initial temporal nodes. Assembling is required to make the edge probability estimated from the whole graph. The complete learning process, including encoding, decoding, and assembling, requires a time complexity of $O(n^2\times(T))$. The parallelization happened when encoding from multiple ego-graphs, decoding for temporal edges, assembling all the probability from one node to all other nodes. After parallelization, the computation steps are $O(n\times T)$ when the number of nodes does not exceed the number of threads (e.g., cuda cores in a NVIDIA GPU).

\begin{algorithm}[t!]
\caption{Decoding k-Radius Temporal Ego-Graph}
\label{alg:decoding_egog}
\SetKwFunction{decode}{k-EgoGraphDecoding}
\SetKwFunction{edgeprob}{EdgeProbability}
\SetKwProg{Fn}{Function}{}{}

\Fn{\decode{$\Tilde{G}_{ego}(u^t)$, $\mathbf{h}_{u^t}$, $\mathbf{X}_{ego}$, $k$}}{
    mu $\gets\text{MLP}_{\mu}(\mathbf{X}_{ego})$\;
    sigma $\gets\text{MLP}_{\sigma^2}(\mathbf{X}_{ego})$\;
    noise $\gets$ random.normal($\mathbf{X}_{ego}$.shape)\;
    $\mathbf{Z}\gets \text{mu} + \text{sigma} * \text{noise}$\;
    \Return{\edgeprob{$\Tilde{G}_{ego}(u^t)$, $\mathbf{h}_{u^t}$, $\mathbf{Z}$, $k$}}\;

}
\Fn{\edgeprob{$\Tilde{G}_{ego}(u^t)$, $\mathbf{h}_{u^t}$, $\mathbf{Z}$, $k$}}{
    p, $\mathbf{h}$, problist $\gets\emptyset$\;
    \eIf{k $\ne$ 1}{
        \ForEach{$v^t\in\mathbf{N}(u^t)$}{
            $\mathbf{h}\gets\mathbf{h}_{u^t}+\mathbf{Z}(v^t,:)$\;
            p $\gets$\edgeprob{$\Tilde{G}_{ego}(v^t)$, $\mathbf{h}$, $\mathbf{Z}$, $k-1$}\;
            problist.extend(p)\;
        }
        \Return{problist}\;
    }{
        $\mathbf{h}\gets\mathbf{h}_{u^t}+\mathbf{Z}(u^t,:)$\;
        p $\gets\text{softmax}(\mathbf{h}\times\mathbf{W}_{dec}+\mathbf{b}_{dec})$\;
        problist.insert(p)\;
        \Return{problist}\;
    }
}
\end{algorithm}

\subsection{Optimization Strategy}
\noindent\textbf{Batch Gradient Descending}.
We jointly optimize encoder and decoder's parameters by minimizing the variational lower bound as follows:

\begin{equation}
    \begin{split}
        P(\mathbf{S}_{t=1:T})=\prod_{e_{i,j}\in \mathbf{A}_{t=1:T}}&P(\mathbf{S}_{t=1:T})_{i,j}\\
        \mathcal{L}=-\frac{1}{NT}\sum_{t=1}^{T}\sum_{u=1}^{N}&\textbf{A}_{u^t}\log(P(\mathbf{S}_{t,u}))+\text{KL}(q(\mathbf{Z|X})||p(\mathbf{Z}))\\
    \end{split}
\end{equation}
where $P(\mathbf{S}_{t=1:T})$ denotes the generated score matrix from decoding module and $\text{KL}(\cdot||\cdot)$ is the Kullback-Leibler divergence between two distributions.

\vspace{2mm}
\noindent\textbf{Mini-batch Gradient Descending and Approximate Loss}.
In practice, it is more efficient to use a mini-batch gradient descending to update the model's parameters. Particularly, we optimize the model parameters through mini-batch data, i.e., randomly sampled ego-graphs and corresponding node features, achieve global parameter training, and can train a model with satisfactory generalization and robustness in less time. In addition, our KL-divergence calculation is still carried out on all nodes. Therefore, in our TGAE implementation, we update the parameters with an approximate loss function, which is formulated as follows:
\begin{equation}
    \begin{split}
        \mathcal{L}_{tgae}=-\frac{1}{n_s}\sum_{u^t\in\Tilde{V}_s}&\textbf{A}_{u^t}\log(P(\mathbf{S}_{t,u}))+\text{KL}(q(\mathbf{Z|X})||p(\mathbf{Z}))\\
    \end{split}
\end{equation}
where $\Tilde{V}_s$ denotes the set of sampled initial temporal nodes and $n_s$ denotes the size of $\Tilde{V}_s$. By adjusting the value of $n_s$, we can achieve the trade-off between generating high-quality temporal graphs and fast model training.

\subsection{Model Variants}
\noindent\textbf{Ego-Graph Sampling Variant.}
Our model can be generalized to random walk-based variants, only by reducing the neighbor threshold $th$ to less than 2 in Algorithm~\ref{alg:sampling_egog}. 
In this case, the ego-graph obtained by our temporal ego-graph sampling strategy is a chain structure, that is, a temporal random walk on the temporal graph. 
In this variant, we fix the whole architecture of TGAE so that it is consistent with the full version, except for the threshold of ego-graph neighbor sampling.

\vspace{1mm}
\noindent\textbf{Initial Node Sampling Variant.}
In addition to ego-graph sampling strategy, our initial node sampling strategy can also be modified to a uniform distribution based sampling strategy. Under this node sampling strategy, our model learns to reconstruct every edge in the temporal graph without bias. In this variant, only the initial node sampling strategy is different from TGAE, and the rest are consistent with the proposed version.

\vspace{1mm}
\noindent\textbf{Non-probablistic Variant.}
We also propose a non-probabilistic variant derived from full TGAE, in which the ego-graph sampling and temporal graph attention encoding are consistent with the full version. 
We modify the decoder of the full TGAE model to non-probabilistic version, which is formulated as follows:

\begin{equation}
    \begin{split}
    \mathbf{Z}\gets\text{MLP}_{\mu}(\mathbf{X}_{ego})
    \end{split}
\end{equation}
where $\mathbf{X}_{ego}$ denotes the input features of the sampled central temporal nodes. Then, the calculation of the approximate loss is modified to fit this variant, which is formulated as follows:

\begin{equation}
    \begin{split}
        \mathcal{L}_{tgae}=-\frac{1}{n_s}\sum_{u^t\in\Tilde{V}_s}&\textbf{A}_{u^t}\log(P(\mathbf{S}_{t,u}))\\
    \end{split}
\end{equation}

After model training and parameter optimization, the temporal graph generation process of the non-probabilistic variant is consistent with the full TGAE.

\begin{table}
  \caption{Statistics of the network data sets.}
  \label{tab:datasets}
  \tabcolsep 9pt
  \begin{tabular}{lccc}
    \hline
    \textbf{Network} & \textbf{\#Nodes} & \textbf{\#Edges} & \textbf{\#Timestamps} \\
    \hline \hline
    DBLP & 1,909 & 8,237 & 15 \\
    \hline
    EMAIL & 986 & 332,334 & 805 \\
    \hline
    MSG & 1,899 & 20,296 & 195 \\
    \hline
    BITCOIN-A & 3,783 & 24,186 & 1,902 \\
    \hline
    BITCOIN-O & 5,881 & 35,592 & 1,904 \\
    \hline
    MATH & 24,818 & 506,550 & 79 \\
    \hline
    UBUNTU & 159,316 & 964,437 & 88 \\
  \hline
\end{tabular}
\end{table}

\begin{table}
  \caption{Graph statistics for measuring network properties.}\vspace{-0pt}
  \label{tab:metrics}
  \begin{tabular}{lcm{2.2cm}}
    \hline
    \textbf{Metric Name} & \textbf{Computation} & \textbf{Description} \\
    \hline \hline
    Mean Degree & $\mathbb{E}[d(v)]$ & Mean degree of nodes. \\
    \hline
    Claw Count & $\sum_{v \in V}\binom{d(v)}{3}$ & \# Claws of the graph. \\
    \hline
    Wedge Count & $\sum_{v \in V}\binom{d(v)}{2}$ & \# Wedges of the graph. \\
    \hline
    Triangle Count & $\frac{trace(A^3)}{6}$ & \# Triangles of the graph. \\
    \hline
    LCC & $max_{f\in F}{||f||}$ & Size of the largest connected component. \\
    \hline
    PLE & $1+n(\sum_{v\in V}{log{(\frac{d(u)}{d_{min}})}})^{-1}$ & Exponent of power-law distribution. \\
    \hline
    N-Component & $|F|$ & \# connected components. \\
  \hline
\end{tabular}
\vspace{-0pt}
\end{table}

\begin{table*}[!h]
\caption{Median score $f_{med}(\cdot)$ comparison with seven metrics. (Smaller metric values indicate better performance)}\vspace{-0pt}
\label{tab:median}
\tabcolsep 4pt
\begin{tabular}{ccccccccccccc}
\toprule
\textbf{Dataset}           & \textbf{Metric} & \textbf{TGAE} & \textbf{TIGGER} & \textbf{DYMOND} & \textbf{TGGAN}  & \textbf{TagGen}  & \textbf{NetGAN}  & \textbf{E-R}      & \textbf{B-A}      & \textbf{VGAE}    & \textbf{Graphite} & \textbf{SBMGNN} \\ \midrule 
\multirow{7}{*}{DBLP}      & Mean Degree     & 2.41E-3 & 3.54E-3 & 2.98E-3 & 3.25E-3 & \textbf{7.46E-4}          & 4.16E-3          & 5.52E-3          & 1.23E-1          & 1.79E-3          & 1.79E-3           & 1.79E-3         \\
                           & LCC             & \textbf{2.61E-3} & 2.75E-3 & 2.71E-3 & 2.77E-3 & 2.78E-3 & 3.35E-1          & 7.27E-1          & 9.11E-2          & 5.11E-1          & 5.40E-1           & 4.62E-1         \\
                           & Wedge Count     & \textbf{4.15E-3} & 3.08E-2 & 2.31E-2 & 5.38E-1 & 7.14E-1          & 5.05E-1          & 5.07E-1          & 3.74E-1          & 1.81E+0          & 2.15E+0           & 2.39E+0         \\
                           & Claw Count      & \textbf{7.29E-3} & 2.64E-2 & 1.35E-2 & 2.98E+0 & 3.02E+0          & 9.27E-1          & 8.78E-1          & 4.52E+0          & 8.78E+0          & 1.21E+1           & 1.42E+1         \\
                           & Triangle Count  & \textbf{4.79E-3} & 7.85E-2 & 3.77E-2 & 5.33E-1 & 5.44E-1          & 8.83E-1          & 9.94E-1          & 8.24E-1          & 9.27E+0          & 9.21E+0           & 8.66E+0         \\
                           & PLE             & \textbf{1.73E-3} & 3.34E-2 & 9.15E-3 & 1.78E-1 & 1.79E-1          & 2.24E-1          & 1.65E-1          & 8.45E-2          & 4.01E-1          & 4.65E-1           & 4.25E-1         \\
                           & N-Components    & \textbf{3.05E-3} & 3.07E-3 & 3.11E-3 & 3.39E-3 & 3.51E-3          & 2.13E-1          & 8.36E-1          & 5.06E-2          & 5.07E-1          & 5.49E-1           & 4.83E-1         \\ \midrule
\multirow{7}{*}{MATH}      & Mean Degree     & \textbf{2.69E-2}    & 1.05E-1 & OOM & OOM       & OOM              & 2.13E-1          & 2.29E-1          & 3.24E-2 & 2.39E-1          & 2.44E-1           & 3.72E-2         \\
                           & LCC             & 8.72E-2 & 9.31E-2 & OOM & OOM   & OOM              & \textbf{2.99E-2}          & 8.83E-1          & 1.24E-1          & 5.56E-1          & 5.30E-1           & 3.73E-1         \\
                           & Wedge Count     & \textbf{1.05E-1} & 2.37E-1 & OOM & OOM   & OOM              & 2.42E-1          & 9.27E-1          & 3.15E-1          & 6.87E-1          & 7.50E-1           & 1.77E+0         \\
                           & Claw Count      & \textbf{2.59E-1} & 3.75E-1 & OOM & OOM   & OOM              & 4.96E-1          & 9.99E-1          & 4.86E-1          & 2.95E+0          & 3.43E+0           & 8.13E+0         \\
                           & Triangle Count  & \textbf{9.79E-2} & 8.78E-1 & OOM & OOM   & OOM              & 2.34E+0          & 1.00E+0          & 5.84E-1          & 1.74E+0          & 1.66E+0           & 2.24E+0         \\
                           & PLE             & \textbf{2.41E-2} & 9.36E-1 & OOM & OOM   & OOM              & 1.11E+0          & 2.35E-1          & 7.81E-2          & 5.74E-1          & 5.40E-1           & 2.49E-1         \\
                           & N-Components    & \textbf{3.15E-2} & 4.66E-2 & OOM & OOM   & OOM              & 3.50E-2          & 1.00E+0          & 1.35E-1          & 5.90E-1          & 5.61E-1           & 3.96E-1         \\ \midrule
\multirow{7}{*}{UBUNTU}    & Mean Degree     & \textbf{9.73E-2} & OOM & OOM & OOM   & OOM              & OOM              & 2.32E+1          & 5.29E-1          & OOM              & OOM               & OOM             \\
                           & LCC             & \textbf{1.32E-1} & OOM & OOM & OOM   & OOM              & OOM              & 3.71E+0          & 2.98E+0          & OOM              & OOM               & OOM             \\
                           & Wedge Count     & \textbf{3.16E-1} & OOM & OOM & OOM   & OOM              & OOM              & 1.45E+1          & 9.76E-1          & OOM              & OOM               & OOM             \\
                           & Claw Count      & 5.60E-1          & OOM & OOM & OOM   & OOM              & OOM              & \textbf{3.01E-1} & 9.96E-1          & OOM              & OOM               & OOM             \\
                           & Triangle Count  & \textbf{1.21E-1} & OOM & OOM & OOM   & OOM              & OOM              & 5.05E-1          & 1.00E+0          & OOM              & OOM               & OOM             \\
                           & PLE             & \textbf{8.52E-2} & OOM & OOM & OOM   & OOM              & OOM              & 7.33E-1          & 5.31E-1          & OOM              & OOM               & OOM             \\
                           & N-Components    & \textbf{2.64E-2} & OOM & OOM & OOM   & OOM              & OOM              & 1.00E+0          & 8.55E-1          & OOM              & OOM               & OOM             \\ \bottomrule 
\end{tabular}
\end{table*}

\begin{table*}[!h]
\caption{Average score $f_{avg}(\cdot)$ comparison with seven metrics. (Smaller metric values indicate better performance)}\vspace{-0pt}
\label{tab:average}
\tabcolsep 4pt
\begin{tabular}{ccccccccccccc}
\toprule
\textbf{Dataset}           & \textbf{Metric} & \textbf{TGAE} & \textbf{TIGGER} & \textbf{DYMOND} & \textbf{TGGAN} & \textbf{TagGen}  & \textbf{NetGAN}  & \textbf{E-R} & \textbf{B-A} & \textbf{VGAE}    & \textbf{Graphite} & \textbf{SBMGNN} \\ \midrule
\multirow{8}{*}{DBLP}      & Mean Degree     & 2.33E-3  & 3.41E-3 & 2.78E-3 & 3.68E-3 & \textbf{1.31E-3} & 3.83E-3          & 2.12E-2     & 1.08E-1     & 8.93E-3          & 8.76E-3           & 8.92E-3         \\
                           & LCC             & \textbf{4.81E-3} & 4.37E-2 & 8.76E-3 & 7.83E-2 & 8.62E-2          & 6.57E-1          & 6.23E-1     & 1.93E-1     & 5.09E-1          & 5.04E-1           & 4.08E-1         \\
                           & Wedge Count     & \textbf{7.46E-3} & 6.81E-1 & 3.06E-2 & 7.25E-1 & 9.88E-1          & 5.63E-1          & 4.76E-1     & 3.50E-1     & 1.92E+0          & 2.11E+0           & 2.36E+0         \\
                           & Claw Count      & \textbf{1.14E-2} & 2.98E-1 & 5.23E-2 & 3.22E+0 & 5.21E+0          & 1.26E+0          & 8.55E-1     & 4.63E+0     & 1.11E+1          & 1.24E+1           & 1.69E+1         \\
                           & Triangle Count  & \textbf{7.38E-3} & 3.84E-1 & 2.95E-2 & 5.24E-1 & 6.86E-1          & 7.51E-1          & 9.92E-1     & 8.16E-1     & 8.99E+0          & 9.61E+0           & 9.20E+0         \\
                           & PLE             & \textbf{2.71E-3} & 1.75E-1 & 3.64E-2 & 2.53E-1 & 2.50E-1          & 2.24E-1          & 1.83E-1     & 8.46E-2     & 3.98E-1          & 4.32E-1           & 4.00E-1         \\
                           & N-Components    & \textbf{3.07E-3} & 3.77E-2 & 9.47E-3 & 4.20E-2 & 4.64E-2          & 2.29E-1          & 6.19E-1     & 1.02E-1     & 5.95E+0          & 6.27E+0           & 5.93E+0         \\ \midrule
\multirow{7}{*}{MATH}      & Mean Degree     & \textbf{2.64E-2} & 6.39E-2 & OOM & OOM & OOM              & 1.97E-1          & 2.08E-1     & 3.81E-2     & 2.05E-1          & 2.07E-1           & 3.97E-2         \\
                           & LCC             & 8.08E-2          & 7.01E-1 & OOM & OOM & OOM              & \textbf{3.15E-2} & 1.49E+0     & 2.41E-1     & 5.41E-1          & 5.16E-1           & 3.68E-1         \\
                           & Wedge Count     & \textbf{1.24E-1} & 1.39E-1 & OOM & OOM & OOM              & 2.57E-1          & 9.30E-1     & 3.29E-1     & 8.07E-1          & 8.59E-1           & 1.86E+0         \\
                           & Claw Count      & \textbf{2.74E-1} & 2.98E-1 & OOM & OOM & OOM              & 4.97E-1          & 9.99E-1     & 5.00E-1     & 3.45E+0          & 3.89E+0           & 8.50E+0         \\
                           & Triangle Count  & \textbf{1.20E-1} & 4.18E-1 & OOM & OOM & OOM              & 2.20E+0          & 1.00E+0     & 6.35E-1     & 1.88E+0          & 1.81E+0           & 2.25E+0         \\
                           & PLE             & \textbf{2.43E-2} & 8.31E-2 & OOM & OOM & OOM              & 9.66E-1          & 2.98E-1     & 1.09E-1     & 5.39E-1          & 5.11E-1           & 2.34E-1         \\
                           & N-Components    & \textbf{9.39E-2} & 1.12E-1 & OOM & OOM & OOM              & 1.34E-1          & 9.45E-1     & 3.19E-1     & 4.73E+0          & 4.52E+0           & 3.33E+0         \\  \midrule
\multirow{7}{*}{UBUNTU}    & Mean Degree     & \textbf{7.41E-2} & OOM & OOM & OOM & OOM              & OOM              & 2.03E+1     & 8.89E-1     & OOM              & OOM               & OOM             \\
                           & LCC             & \textbf{2.10E-1} & OOM & OOM & OOM & OOM              & OOM              & 6.94E+3     & 6.02E+3     & OOM              & OOM               & OOM             \\
                           & Wedge Count     & \textbf{3.06E-1} & OOM & OOM & OOM & OOM              & OOM              & 5.07E+4     & 2.34E+4     & OOM              & OOM               & OOM             \\
                           & Claw Count      & \textbf{5.14E-1} & OOM & OOM & OOM & OOM              & OOM              & 3.10E+5     & 3.29E+6     & OOM              & OOM               & OOM             \\
                           & Triangle Count  & \textbf{1.01E-1} & OOM & OOM & OOM & OOM              & OOM              & 4.73E-1     & 7.95E-1     & OOM              & OOM               & OOM             \\
                           & PLE             & \textbf{1.29E-1} & OOM & OOM & OOM & OOM              & OOM              & 6.51E-1     & 5.37E-1     & OOM              & OOM               & OOM             \\
                           & N-Components    & \textbf{8.40E-2} & OOM & OOM & OOM & OOM              & OOM              & 9.97E-1     & 8.25E-1     & OOM              & OOM               & OOM             \\   \bottomrule
\end{tabular}
\end{table*}

\subsection{Temporal Graph Assembling and Generation}\label{subsec:assem}
After the training process, we first generate all the ego-graphs to assemble the score matrix $S$. Specifically, the algorithm~\ref{alg:decoding_egog} shows the procedure of decoding the edge probabilities of a $k$-radius ego-graph for each temporal node $u^t$, where the $\mathbf{W}_{dec}\in\mathbb{R}^{d_{in}\times n}$ and $\mathbf{b}_{dec}\in\mathbb{R}^{n}$ are the output parameters for decoding process.
When the categorical distribution of the $i-1$-order neighbors of $u^t$ is generated, we generate $i$-order neighbors' edge probabilities. The process finishes when all the $k$-radius ego-graph's generative distributions are generated. Assuming that $\mathbf{H}\in\mathbb{R}^{nT\times d_{enc}}$ contains all hidden variables of all temporal nodes, we generate score matrix $\mathbf{S}_{t=1:T}$ by averaging all the edge probabilities generated by the ego-graphs.
Then, we take score matrix as the parameters of the categorical distribution of each temporal edge with $p(t,u,v)=\frac{\mathbf{S}_{t,u,v}}{\sum_{i\in\mathbf{N}(u^t)}\mathbf{S}_{t,u,i}}$.
Then we sample the corresponding temporal edges for each temporal node without replacement, which is formulated as $\mathbf{A}_{u^t}'\sim Cat(\prod_{i\in\mathbf{N}(u^t)}p(t,u,i))$, where $Cat$ denotes categorical distribution. The generation process finishes when the generated temporal graph's edge amount meets the one observed graph.





\section{Experiments}
\label{sec:exprm}

In this section, we describe the extensive experiments for evaluating the effectiveness of our proposed method. We first describe the experiment setup. Then, present the experimental results of temporal graph auto-encoder compared with other baselines, which is the main task of this paper. After that, the efficiency and scalability of the proposed method are tested. Finally, we report the ablation study and parameter sensitivity experiments.


\begin{table*}[!h]
\caption{Maximum mean discrepancy of instance counts of all 2- and 3-node, 3-edge \bm{$\delta$}-temporal motifs between raw and generated temporal networks (\bm{$\sigma$} refers to the sigma value for Gaussian kernel)}
\label{tab:motifs}
\tabcolsep 4pt
\begin{tabular}{cccccccccccc}
\hline
\textbf{Dataset} & \textbf{TGAE} & \textbf{TIGGER} & \textbf{DYMOND} & \textbf{TGGAN} & \textbf{TagGen} & \textbf{NetGAN} & \textbf{E-R} & \textbf{B-A} & \textbf{VGAE} & \textbf{Graphite} & \textbf{SBMGNN} \\ \hline \hline
DBLP             & \textbf{2.65E-5} & 9.68E-4 & 1.25E-4 & 2.08E-2 & 2.31E-2         & 2.21E-1         & 6.43E-2     & 1.08E+0     & 1.34E+0       & 1.95E+0           & 1.99E+0         \\
MSG              & \textbf{2.27E-5} & 2.12E-4 & 3.77E-5 & 9.81E-3 & 1.09E-2         & 1.85E-2         & 1.83E-2     & 1.17E+0     & 1.98E+0       & 1.99E+0           & 1.65E+0         \\
BITCOIN-A        & \textbf{1.12E-6} & 2.76E-5 & OOM & OOM & OOM             & OOM             & 1.90E+0     & 2.00E+0     & 3.88E-1       & 5.39E-1           & 1.08E-1         \\
BITCOIN-O        & \textbf{5.49E-6} & 3.06E-5 & OOM & OOM & OOM             & OOM             & 1.80E+0     & 2.00E+0     & 1.82E+0       & 1.98E+0           & 5.22E-1         \\
EMAIL            & \textbf{2.12E-2} & 7.65E-2 & 3.27E-2 & OOM & OOM             & 9.78E-2             & 9.74E-1     & 1.95E+0     & 1.95E+0       & 1.07E+0           & 1.74E+0         \\
MATH             & \textbf{7.86E-4} & 2.14E-3 & OOM & OOM & OOM             & 5.11E-3         & 6.59E-3     & 2.74E-3     & 2.00E+0       & 1.89E+0           & 1.94E+0         \\
UBUNTU           & \textbf{1.27E-3} & OOM & OOM & OOM & OOM             & OOM             & 1.52E+0     & 2.00E+0     & OOM           & OOM               & OOM             \\ \hline
\end{tabular}
\end{table*}

\subsection{Experiment Settings}
\label{sec:expset}

We introduce the experimental datasets, comparison methods, metrics, and parameter settings in this subsection.

\vspace{1mm}
\noindent \textbf{Datasets.} We evaluate our temporal graph auto-encoder on seven real temporal networks. Specifically, DBLP \cite{zhou2015rare} is a citation network that contains bibliographic information of the publications in IEEE Visualization Conference from 1990 to 2015; MSG \cite{panzarasa2009patterns} and EMAIL \cite{paranjape2017motifs} are two communication networks, where a single edge represents a message/email sent from one person to another at a certain timestamps; BITCOIN-A and BITCOIN-O \cite{kumar2016edge, kumar2018rev2} are two who-trusts-whom networks where people trade with bitcoins on Bitcoin Alpha and OTC platforms; MATH and UBUNTU \cite{paranjape2017motifs} are temporal networks of interactions on the stack exchange web sites Math Overflow and Ask Ubuntu. The statistics of datasets are summarized in Table~\ref{tab:datasets}.

\vspace{1mm}
\noindent \textbf{Compared methods.} We compare TGAE with two state-of-the-art temporal graph generative models (TIGGER \cite{Gupta_Manchanda_Bedathur_Ranu_2022_TIGGER} and DYMOND \cite{WWW21DYMOND}), three GAN-based graph generative models (TGGAN \cite{zhang2021tggan}, TagGen \cite{zhou2020data}, and NetGAN \cite{bojchevski2018netgan}),  two simple model-based generative models (E-R \cite{erdHos1960evolution} and B-A \cite{albert2002statistical}), and three auto-encoder-based generative models (VGAE \cite{kipf2016variational}, Graphite \cite{grover2019graphite}, and SBMGNN \cite{mehta2019stochastic}). Note that NetGAN, simple model-based, and autoencoder-based models are not designed for temporal graph generation. To generate temporal networks, we separately generate snapshots of the temporal graph at each timestamp.

\begin{figure*}[t!]
    \centering
    \includegraphics[width=0.9\linewidth]{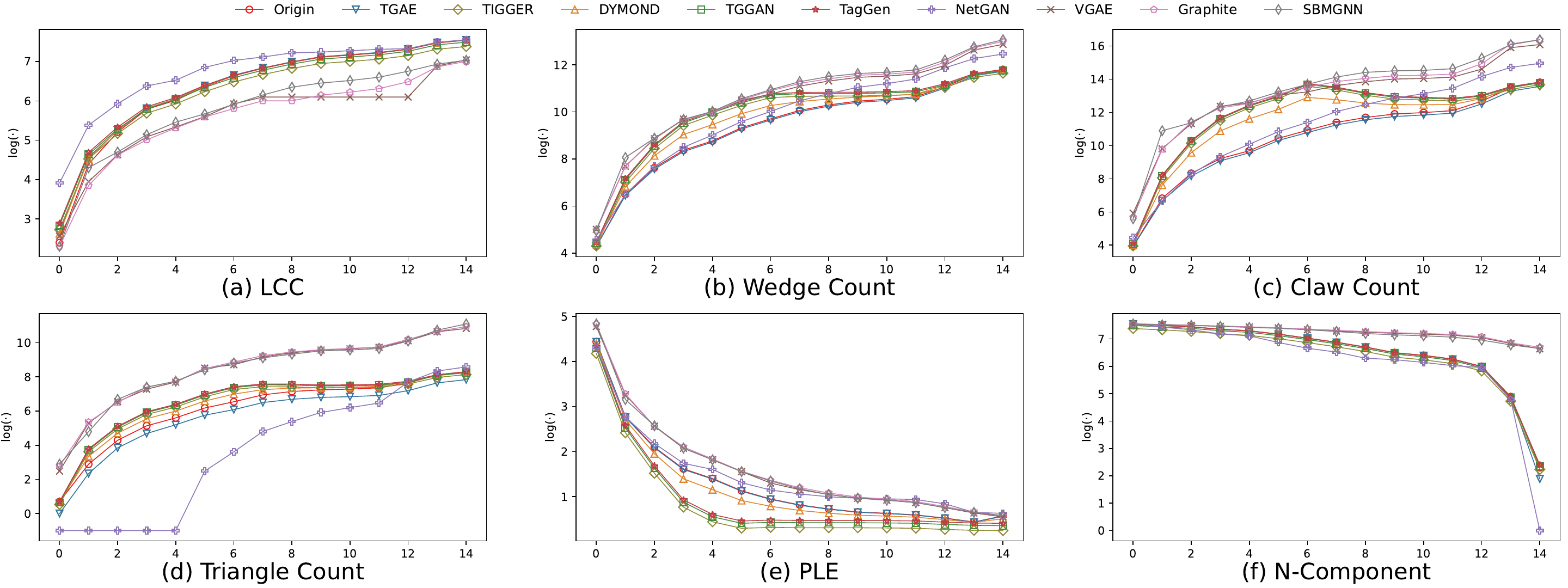}
    \caption{The comparison results on the seven evaluation metrics across 15 timestamps in DBLP data set. Best viewed in color. The algorithm better fitting the curve of the original graph (colored in blue) is better.}
    \label{fig:DBLP_compare}
\end{figure*}

\vspace{1mm}
\noindent \textbf{Evaluation metrics.} We collected several popular evaluating metrics to measure the difference between the original temporal graph and the generated graph. The graph statistics for measuring graph properties are summarized in Table \ref{tab:metrics}. As all of these metrics are designed for static graphs, we follow the practice of TagGen~\cite{zhou2020data}, who generalized the aforementioned metrics to the dynamic setting by calculating mean and median value of the metrics among all timestamps. Specifically, given a metric $f_m(\cdot)$, the real graph $\widetilde{G}$, and the synthetic one $\widetilde{G'}$ , we construct a sequence of snapshots $\widetilde{S^t}\ (\widetilde{S'}^t),\ t=1,\dots,T$, of $\widetilde{G}\ (\widetilde{G'})$ by aggregating edges from the initial timestamp to the current timestamp $t$. Then, we measure the average/median difference (in percentage) of the given metric $f_m(\cdot)$ between two graphs as follows:
\begin{equation}
\begin{aligned}
    f_{avg}(\widetilde{G},\ \widetilde{G'},\ f_m) =& Mean_{t=1:T}(|\frac{f_m(\widetilde{S^t})-f_m(\widetilde{S'}^t)}{f_m(\widetilde{S^t})}|) \\
    f_{med}(\widetilde{G},\ \widetilde{G'},\ f_m) =& Median_{t=1:T}(|\frac{f_m(\widetilde{S^t})-f_m(\widetilde{S'}^t)}{f_m(\widetilde{S^t})}|)
\end{aligned}
\end{equation}

\vspace{1mm}
\noindent \textbf{Parameter settings.} 
As to baseline methods, we use the best parameter settings and GPU-accelerated version (if applicable) given by the original authors. Our proposed TGAE and evaluating metrics are implemented through Python-3.7, PyTorch-1.8, and CUDA-11.1 in our experiments. The experiments are operated on a machine with Intel(R) Xeon(R) Gold 5220 CPU @ 2.20GHz, 62 GB RAM, and NVIDIA Tesla V100 with 32 GB memory. We use one CPU core and one GPU for every algorithm.

\subsection{Temporal Graph Generation}
We compare our proposed TGAE with ten baseline models across seven temporal graph datasets. For the static methods, we apply them to generate one static graph at each timestamp and construct a series of graph snapshots by aggregating all static graphs. The results of seven evaluating metrics in the form of $f_{avg}(\cdot)$ and $f_{med}(\cdot)$ are shown in Tables \ref{tab:average} and \ref{tab:median}.


\noindent\textbf{Evaluation with graph statistics}
As shown in Tables~\ref{tab:average} and \ref{tab:median}, TGAE outperforms all the baseline methods in at least six of seven evaluating metrics. As to the DBLP dataset, the state-of-the-art baseline DYMOND achieves the second-best performance. Besides, TagGen achieves the best performance on \textit{mean degree} measurement. As to MATH dataset, TIGGER achieves the second-best performance. According to the seventh and eighth columns, simple model-based generative methods (i.e., E-R and B-A) has the worst generative performance on temporal graph generation. According to the sixth column and the last three columns, we can see that static graph generative methods (i.e., NetGAN, VGAE, Graphite, and SBMGNN) are consistently worse than temporal graph generative methods.
TGAE significantly outperforms TIGGER, DYMOND, TGGAN, and TagGen with all metrics except a slightly worse with Mean Degree, which shows that TGAE is better at capturing most graph properties. TGAE also outperforms other methods in the other datasets (e.g., MSG dataset). Due to space limits, we put the representative results in this manuscript. Please note that most of the learning-based methods cannot simulate large temporal graphs (e.g., the UBUNTU dataset, containing about 14 million temporal nodes) due to their high requirements for GPU memory usage. Our proposed TGAE can simulate these large temporal graphs with affordable time and space consumption. We used datasets of moderate size in this experiment to ensure that all baseline methods could run without encountering out-of-memory (OOM) issues. When we attempted to use larger datasets for these comparisons, the other methods failed due to OOM errors, preventing a fair assessment of generation quality.

\begin{table}[!h]
\caption{Results of ablation study on TGAE and its variants. (Smaller metric values indicate better performance)}
\label{tab:ablation}
\tabcolsep 3pt
\begin{tabular}{ccccccc}
\toprule
\textbf{Dataset}           & \textbf{Metric} & \textbf{TGAE} & \textbf{TGAE-g} & \textbf{TGAE-t} & \textbf{TGAE-n}  & \textbf{TGAE-p} \\ \midrule
\multirow{2}{*}{MSG}      & Degree     & \textbf{1.61E-2}  & 3.66E-2 & 1.65E-2 & 1.73E-2 & 1.85E-2       \\
                           & Motif    & \textbf{2.27E-5} & 8.14E-5 & 2.95E-5 & 4.67E-5     & 4.93E-5    \\ \midrule
\multirow{2}{*}{BITCOIN-A}      & Degree     & \textbf{5.18E-3} & 1.27E-2 & 5.89E-3 & 7.33E-3 & 7.91E-3       \\
                           & Motif    & \textbf{1.12E-6} & 4.33E-6 & 1.78E-6 & 2.98E-6    & 2.35E-6       \\  \midrule
\multirow{2}{*}{BITCOIN-O}    & Degree     & \textbf{1.11E-2} & 2.33E-2 & 1.27E-2 & 1.65E-2      & 1.73E-2       \\
                           & Motif    & \textbf{5.49E-6} & 2.10E-5 & 6.81E-6 & 1.09E-5    & 1.13E-5         \\   \bottomrule
\end{tabular}
\end{table}

\subsection{Temporal Attribute Preservation}
We further study the generative performance from detailed comparison based on preserving temporal attributes, such as temporal motifs and temporal tendency.

Temporal motifs are recurring subgraph patterns over time in a temporal graph~\cite{paranjape2017motifs}. By evaluating our model using temporal motifs, we can assess how well our model captures and reproduces these fundamental patterns in the generated graphs. Temporal tendency refers to the propensity of a graph's structure to evolve over time~\cite{zhou2020data}. By evaluating our model using temporal tendency, we can assess how well our model captures and reproduces these dynamic changes in the generated graphs. 

If our model can accurately generate graphs with similar temporal motif distributions as the original graph, it indicates that our model has successfully learned the underlying structural and temporal patterns in the data. And if our model can accurately generate graphs with similar temporal tendencies as the original graph, it indicates that our model has successfully learned the underlying dynamic behaviors in the data.

\vspace{1mm}
\noindent \textbf{Temporal Motif Preservation.}
To evaluate the capacity of preserving temporal pattern information in the observed data, we also count the instances of all 2- and 3-node, 3-edge temporal motifs~\cite{paranjape2017motifs} and calculate the motif distributions \textit{maximum mean discrepancy}~\cite{you2018graphrnn} between the generated graphs and raw temporal graphs. The results on motif distribution are shown in Table \ref{tab:motifs}.

According to the first column of Table~\ref{tab:motifs}, TGAE achieves the best performance in preserving the motif distribution in simulated temporal networks, which demonstrates its ability to capture both temporal and topological information. According to the triangle count row of Tables~\ref{tab:median} and \ref{tab:average}, we also find that the result of preserving motif distribution shows a similar trend to the triangle count. The results indicates that our proposed TGAE can simulate temporal graphs with similar motif distribution compared with observed graphs.

\vspace{1mm}
\noindent \textbf{Temporal Tendency Visualization.}
To visualize the temporal tendency and show the similarity of simulated graphs, we experiment with the DBLP dataset and measure the statistics of observed graphs and generated graphs from all algorithms in each timestamp. By putting results together with the original graph, we can explore more information on the variation tendency of different methods on 15 timestamps in the DBLP dataset. The experimental results are reported in Figure \ref{fig:DBLP_compare}, where the X-axis represents the timestamp, and the Y-axis represents the value of a metric. According to Figurea~\ref{fig:DBLP_compare} (a) and (f), most of all the methods can have similar number of connected components compared to observed graphs. According to Figures~\ref{fig:DBLP_compare} (b) and (c), TGAE (colored in blue) constantly performs better than the baseline methods as it better fits the triangle and claw count variation trends of the original graph (colored in red).  Significantly worse performance of simple model-based algorithms (e.g., E-R) in motif metrics (e.g., Triangle Count) proves their extremely weak expressive power. Our model has surpassed TIGGER, DYMOND, TGGAN, and TagGen in almost all measurements. The results demonstrate that TGAE is the best learning-based temporal graph generative model  in terms of generative quality.

\begin{figure*}[t!]
    \centering
    \includegraphics[width=0.9\linewidth]{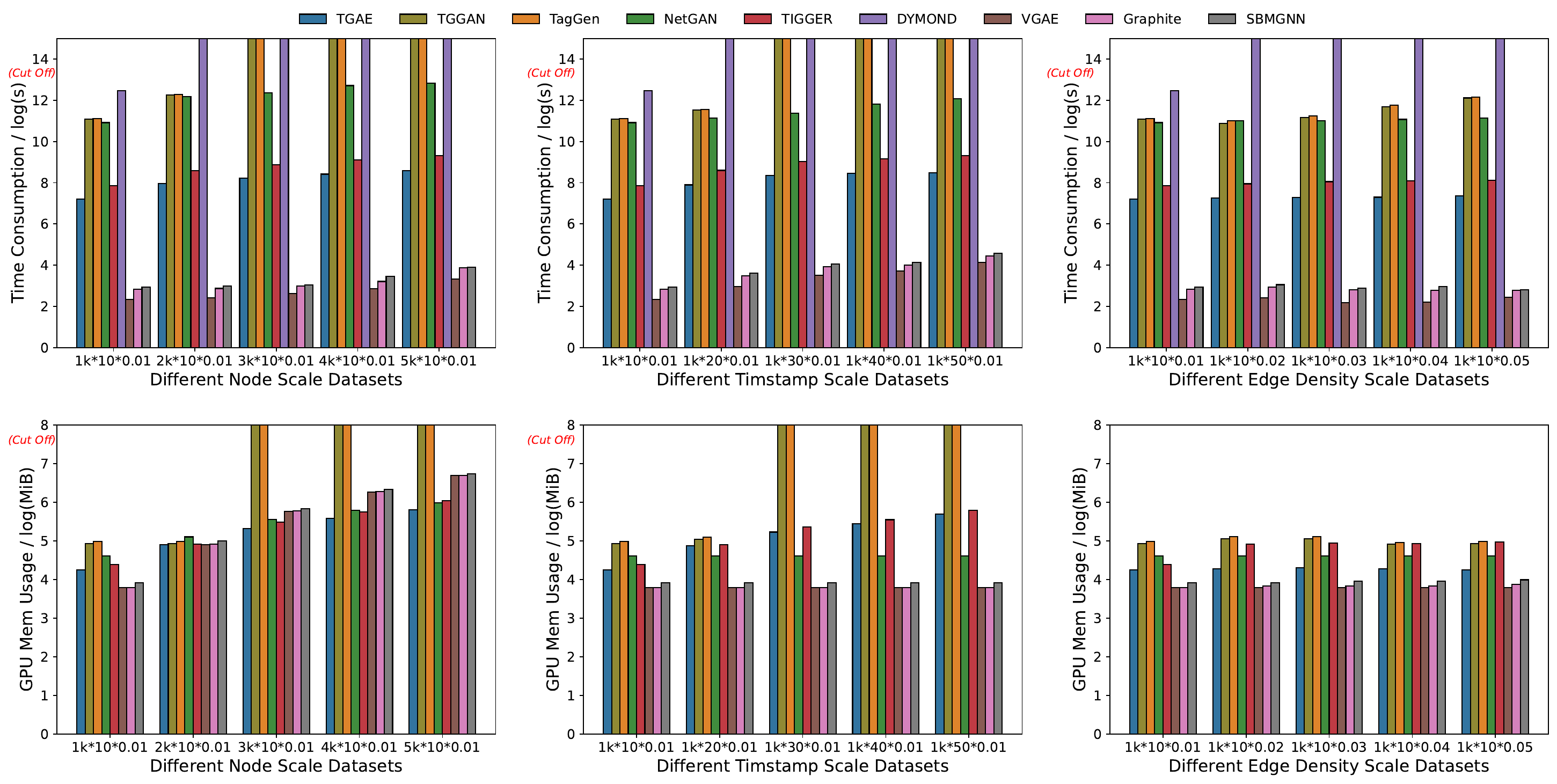}
    \caption{The comparison results on the time consumption and GPU memory usage in data sets designed for scalability test. The x-axis label implies the complexity of input temporal graphs in the form of Number of Nodes $*$ Timestamps $*$ Edge Density.}
    \label{fig:scalability_test}
\end{figure*}

\subsection{Ablation Study.}
To validate the effectiveness of each component and our proposed sampling strategy, we report the ablation study results in Table~\ref{tab:ablation}. TGAE-g denotes the variant that the ego-graph sampling strategy is blocked. TGAE-t denotes the variant that sampling strategy does not truncate the number of neighbor nodes. TGAE-n denotes the node sampling strategy is changed to uniform sampling, that is, blocking the re-weighting strategies. TGAE-p denotes the non-probabilistic variant. According to the first two columns of Table~\ref{tab:ablation}, we can see that if we use the random walk instead of ego-graphs to model the temporal graph, the generative performance degrades significantly. The results in the third column show competitive performance with the complete version, which demonstrates that the truncating process gives a better efficiency w.o. reducing generative ability. The other two variants show similar observations that each component is effective. Due to the paper length limit, we use two representative evaluating metrics and datasets in the ablation experiment, the results are similar to observations from other metrics and datasets. Therefore, the results of the ablation study demonstrate that all the included sampling strategies and components are effective.

\subsection{Scalability and efficiency}
\label{sec:efficiency}
We also evaluated the scalability and efficiency of our model and the baseline methods. The first row of Figure \ref{fig:scalability_test} reports the time consumption of inferring a new graph, whose independent variables are the number of nodes, timestamps, and edge density respectively, while the second row reports the peak memory usage. Note that the B-A and E-R methods are not compared in GPU memory usage, because they are not implemented with the deep learning-based approach.

As the number of sizes (e.g., time stamps, nodes, and edge density) increases, simple model-based graph generators (B-A and E-R) have the highest efficiency for generating large temporal networks, incurring minor extra space costs and taking little time. All the learning-based methods (including our proposed TGAE) are more time-consuming than simple model-based methods.

As for learning-based temporal graph generative models, TGAE achieved much better results than DYMOND, TGGAN, and TagGen in terms of time consumption and memory usage. As can be seen in Table \ref{tab:average}, \ref{tab:median}, and Figure \ref{fig:scalability_test}, learning-based methods, including TGGAN, TagGen, and NetGAN, cannot run through most of the datasets due to its high time and memory consumption. As the number of nodes and timestamps increase, TGAE has a linear increase in time consumption and memory usage. Compared with other learning-based temporal graph generative methods, TGAE is the best model choice for efficient and effective temporal graph generation. Compared with all the learning-based baseline methods, TGAE can achieve a good trade-off between simulating quality and efficiency.

\section{Conclusion}
\label{sec:conclusion}
Temporal graph simulation can help to mimic real-life graphs in many applications, including biology, information technology, and social science. However, most of the graph simulation works focus on static graph simulation, ignoring the temporal evolving property of real-life graphs. In this paper, we proposed temporal graph autoencoders (TGAE) to simulate real-life graphs and reproduce the temporal and structural properties from observed graph data. Besides, existing learning-based approaches are limited by their high memory usage and time consumption, especially for sampling random walk-based approaches. Therefore, we propose initial node sampling and ego-graph sampling strategies to achieve efficient graph generative model training. Extensive experiment results on simulating quality and model efficiency show that our proposed TGAE achieves the best generative performance compared with other learning-based baselines. TGAE also achieves a good trade-off between quality and efficiency. In the future, we aim to scale the learning-based approaches to simulate large graphs with billion nodes.

\bibliographystyle{IEEEtran}
\bibliography{reference}

\end{document}